# Building Low-Resource NER Models Using Non-Speaker Annotations

**Tatiana Tsygankova**♮**, Francesca Marini**♮**, Stephen Mayhew**♭**, Dan Roth**♮

♮University of Pennsylvania, Philadelphia, PA, 19104
♭Duolingo, Pittsburgh, PA, 15206
`ttasya@seas.upenn.edu, fmarini@seas.upenn.edu`
`stephen@duolingo.com, danroth@seas.upenn.edu`

## Abstract

In low-resource natural language processing (NLP), the key problems are a lack of target language training data, and a lack of native speakers to create it. Cross-lingual methods have had notable success in addressing these concerns, but in certain common circumstances, such as insufficient pre-training corpora or languages far from the source language, their performance suffers. In this work we propose a complementary approach to building low-resource Named Entity Recognition (NER) models using "non-speaker" (NS) annotations, provided by annotators with no prior experience in the target language. We recruit 30 participants in a carefully controlled annotation experiment with Indonesian, Russian, and Hindi. We show that use of NS annotators produces results that are consistently on par or better than cross-lingual methods built on modern contextual representations, and have the potential to outperform with additional effort. We conclude with observations of common annotation patterns and recommended implementation practices, and motivate how NS annotations can be used in addition to prior methods for improved performance.[1]

## 1 Introduction

Work in low-resource languages is not only academically compelling, breaking from popular use of massive compute power on unlimited English data, but also useful, resulting in improved digital tools for under-resourced communities. Two common strategies for low-resource NLP include (*a*) building cross-lingual models, and (*b*) annotating data in the target language.

Cross-lingual approaches — in which models are trained on some high-resource language, and applied to the target language — have been shown to be surprisingly effective (Wu and Dredze, 2019; Lample and Conneau, 2019). However, in common circumstances, such as when working with languages with insufficient training corpora or those far from the available source languages, cross-lingual methods suffer (Wu and Dredze, 2020; K et al., 2020). Absent sufficient cross-lingual methods, conventional wisdom suggests that only native (or fluent) speakers of a language can provide useful data to train NLP models. But in low-resource scenarios, fluent speakers may not be readily available.

---

[1]For more details, see:
`http://cogcomp.org/page/publication_view/941`

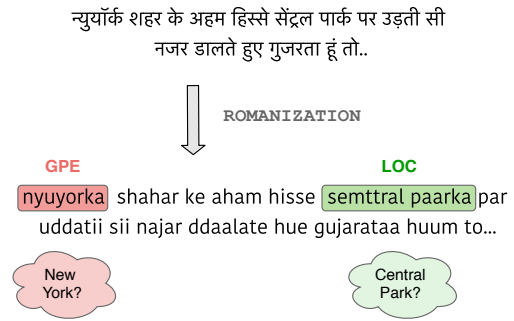

Figure 1: An example of how romanized Hindi text can be annotated without prior language knowledge.

To address this limitation, we hypothesize that the search for annotators can be extended beyond fluent speakers. In this work, we propose an unconventional approach for low-resource named entity recognition (NER) by getting annotations from annotators with no familiarity in the target language, referred to as "non-speaker" (NS) annotation. We posit that annotators are able to use phonetic, syntactic, and even semantic information from their languages of fluency to inform recognition. One example of how phonetic information can be used for NER annotation is shown in Figure 1.

We test our hypothesis in a carefully controlled annotation experiment, comparing the performance of non-speaker (NS) annotators to that of fluent speakers (FS) in Indonesian, Russian, and Hindi.

Our findings are summarized in two key takeaways: (1) non-speaker annotators are able to produce useful annotations despite having no experience annotating or learning the target language; and (2) non-speaker annotations are on par or better than cross-lingual methods built on modern contextual representations. We conclude with observations over factors that can influence NS annotation quality, such as availability of a good romanization system, or presence of capitalization in the target language.

## 2 Related Work

Named Entity Recognition (NER) has been studied for many years (Ratinov and Roth, 2009; Lample et al., 2016; Ma and Hovy, 2016), with most focus on English and a few other European languages (Tjong Kim Sang and De Meulder, 2003).

Recently, there has been growing interest in low-resource NLP, with work in part-of-speech tagging (Plank and Agić, 2018), parsing (Rasooli and Collins, 2017), machine translation (Xia et al., 2019), and other fields. Low-resource NER has seen work using Wikipedia (Tsai et al., 2016), self attention (Xie et al., 2018), and multilingual contextual representations (Wu and Dredze, 2019).

There has been a small amount of work using non-speaker annotations (Mayhew et al., 2019a), but mainly as an application of a technique, falling short of the exhaustive study in this paper.

Several interfaces exist for non-speaker annotations in NER, including TALEN (Mayhew, 2018), which we use, ELISA IE (Lin et al., 2018), and Dragonfly (Costello et al., 2020), which performed small-scale experiments with non-speaker annotators.

A similar approach has been proposed for machine translation (Hermjakob et al., 2018b) and speech recognition (Chen et al., 2016). In the former case (assuming the translation direction is Foreign-to-English), it is often sufficient to translate several of the most important content words, then reconstruct the most likely sentence that uses these. In speech recognition, it is possible to listen to a language one does not speak, and produce a phonetic transcriptions that can be aggregated with others into a reasonable transcription, a process referred to as *mismatched crowdsourcing*.

| Language | Script | Capitalization | Example |
|---|---|---|---|
| Indonesian | Latin | Yes | Amerika |
| Russian | Cyrillic | Yes | Америка |
| Hindi | Devanagari | No | अमेरिका |

Table 1: Factors contributing to language difficulty, with examples of the English word "America."

## 3 Experimental Setup

Our experiment consisted of a series of trials, typically attended by 1–5 participants. Each trial ran for four hours and consisted of three tasks: (1) one-hour instructional training, (2) 20-minute English annotation exercise, and (3) series of five 30-minute sessions annotating documents in the target language.

**Language Selection** We chose three target languages: Indonesian, Russian, and Hindi. These languages were chosen based on availability of gold-annotated data and fluent speakers, and language difficulty.

The constraint of available fluent speakers for annotation, which we use as a point of comparison on non-speaker annotation performance, led us to choose mid- to high-resource languages for evaluation. To read accounts of similar techniques used on true low-resource languages, see the applications section (§4.3).

We define language difficulty as the task-specific difficulty experienced by an English speaker creating NER annotations in the target language. In practice, this difficulty mainly depends on script and capitalization, but may also depend on other factors such as language family and number of English loanwords. Under this task-specific definition and relevant properties summarized in Table 1, Indonesian is identified as the "easiest" language, Russian is "intermediate," and Hindi is the "hardest."

**Participant Selection** In total, there were 30 participants involved in the study, selected largely through a network of friends and acquaintances at the University of Pennsylvania. All participants were uniformly paid $10/hour for their time and were preliminarily screened for language exposure. We chose not to use crowd-sourcing platforms, such as Mechanical Turk, to allow flexibility in administration format and recruitment strategy. The methodology for the study was approved by the Institutional Review Board at the university.

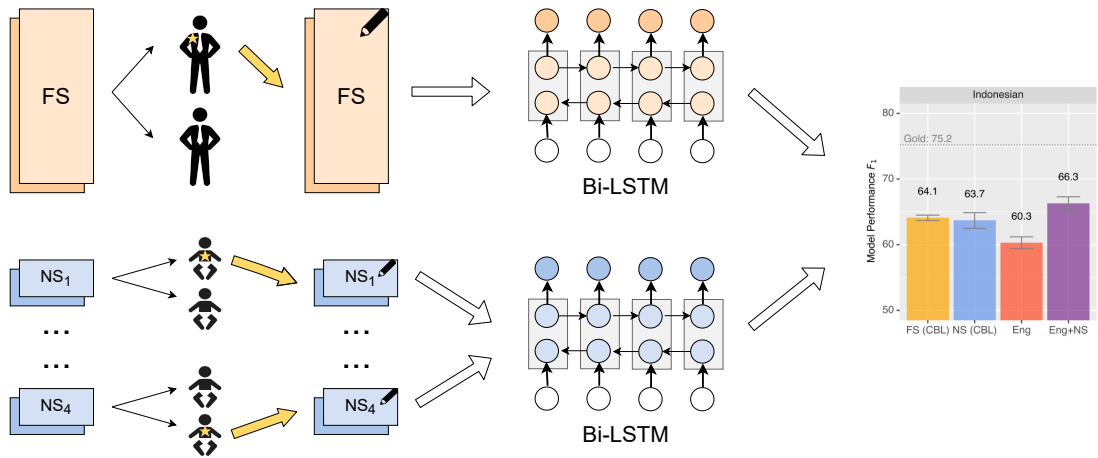

Figure 2: An overview of the data selection process involved in training models on the FS (fluent speaker) and NS (non-speaker) annotations. In each document set, the stars refer to annotators with the higher English exercise score, whose data is used in training. Details on model performance for each language are shown in Figure 3

| Language | Train | Dev | Test |
|---|---|---|---|
| Indonesian | 76K | 18K | 16K |
| Russian | 59K | 16K | 16K |
| Hindi | 72K | 18K | 20K |

Table 2: Size of LORELEI datasets for each language, measured in tokens. Splits were created by the authors.

| Language | FS | NS |
|---|---|---|
| Indonesian | 19K | 38K |
| Russian | 28K | 38K |
| Hindi | 18K | 45K |

Table 3: Size of datasets produced by fluent speaker (FS) and non-speaker (NS) annotators, in tokens.

**Data**  We used gold-annotated NER data from the LORELEI project (Strassel and Tracey, 2016; Tracey et al., 2019). This data uses 4 entity tags: Person, Organization, Location, and Geopolitical Entity. We created splits of these datasets ourselves, statistics of which can be seen in Table 2. These corpora are not parallel.

Accounting for annotation speed differences, FS and NS annotators were given document sets of different sizes to annotate during the same time frame. Each document set used in the experiment was annotated by at least two participants. (visual reference in Figure 2).

**Task 1: Instructional Training**  In total, two instructional documents were used – one providing an overview of the task goals and annotation software, and the other outlining key annotation principles in the form of an interactive annotation guideline quiz. The annotation software used was TALEN (Mayhew, 2018), a tool designed for annotating named entities when the annotators don't speak the target language.

**Task 2: English Annotation Exercise**  After the quiz, participants were asked to annotate English LORELEI data for 20 minutes. The goal of this exercise was both to familiarize the participants with the software interface and provide an indicator of their annotator potential and understanding of the annotation guidelines, used later to filter out low-quality annotators.

**Task 3: Target Language Annotation Sessions**  Participants completed their 2.5 hours of annotation in 5 sessions of 30 minutes each. All FS annotators spent their time annotating documents in their native language, while NS annotators worked with foreign languages that they had no prior exposure to. Given that all of the languages used in the study were high- to mid-resource, annotators were given explicit instructions not to use external model resources such as Google Translate, but were allowed to use internet search to determine the nature of the entities. For Russian and Hindi, which do not use Latin script, we provided uroman (Hermjakob et al., 2018a) romanization, so that the script was not a barrier to successful annotation (Figure 1).

Summary statistics of the annotated documents can be found in Table 3. Note that the larger annotated data size from the NS annotators reflects the fact that there were more NS annotators than FS annotators, a choice we deliberately made.

| | FS | | | NS | | |
|---|---|---|---|---|---|---|
| Language | P | R | $F_1$ | P | R | $F_1$ |
| Indonesian | 80.6 | 75.6 | 78.0 | 59.8 | 55.7 | 57.7 |
| Russian | 69.0 | 67.3 | 68.1 | 57.0 | 45.9 | 50.9 |
| Hindi | 85.5 | 80.4 | 82.9 | 59.8 | 33.4 | 42.8 |

Table 4: Annotation quality of annotations collected from fluent speaker (FS) and non-speaker (NS) annotators against the gold data.

| | Annotation Time (hr) | | | | |
|---|---|---|---|---|---|
| Language | 0.5 | 1 | 1.5 | 2 | 2.5 |
| Indonesian | 55.5 | 54.7 | 57.8 | 57.3 | 57.5 |
| Russian | 42.5 | 47.5 | 47.5 | 49.7 | 50.6 |
| Hindi | 24.5 | 32.3 | 41.1 | 40.3 | 41.8 |

Table 5: Changes in mean annotation quality of non-speaker (NS) annotations over time show an upwards trajectory that steepens with language difficulty.

## 4 Experiments & Analysis

This section describes the analysis done on the gathered FS and NS annotations, through the setup of our models and metrics used and key experimental takeaways.

### 4.1 Models & Metrics

**Two Performance Measures** In this work, we report two distinct $F_1$ performance measures: *Annotation Quality* and *Model Performance*.

*Annotation Quality* refers to the results of participant annotation compared to the existing gold annotations on the same documents. In this evaluation, no model is trained, and we simply calculate the $F_1$ scores by treating NS annotations as predictions themselves (results reported in Tables 3 and 5).

In contrast, *Model Performance* refers to the more traditional NER setup, in which we train a model over obtained annotations, and predict on some held out test set. The following sections outline the results of this performance metric (results reported in Figure 3).

**Data Preparation** To account for random errors, we prioritized recruiting at least two participants to annotate each document set. We then used English exercise scores to choose between the resulting conflicting annotations for the same document sets. A summary of the data selection process is shown in Figure 2.

In order to ensure that documents lacking annotations were considered to be NS annotator mistakes rather than negative training examples, we removed all empty documents from the NS data before training. No other pre-processing was done.

**Machine Learning Models** For all experiments, we used a standard BiLSTM-CRF model (Ma and Hovy, 2016) implemented in AllenNLP (Gardner et al., 2018), and used multilingual BERT embeddings (Devlin et al., 2019), which have been shown to exhibit surprising cross-lingual properties (Wu and Dredze, 2019). For the sake of speed and simplicity, we use BERT embeddings as features, and do not fine-tune the model. For each dataset, we train with 5 random seeds (Reimers and Gurevych, 2017) and report the average.

We recognize that these annotations are missing many entities. Following recent work on partial annotations, we use an iterative method from (Mayhew et al., 2019a) called Constrained Binary Learning (CBL) that detects unmarked tokens likely to be entities and down-weights them in training. Subsequent results reported use this method on all FS and NS annotations.

**Baseline Comparisons** Given that there is little prior work on this subject, it's hard to compare our results against an established baseline. To contextualize our results, we compare NS models against FS models and cross-lingual methods. However, both are imperfect comparisons and should be interpreted with caution.

In our comparison with FS models, the main difficulty is unequal training data size. Our experimental design intentionally left us with more NS annotations than FS annotations (see Table 3). It might be tempting to address this difficulty by balancing data sizes, however constraining the NS annotations to the sizes of the FS data would not give a fair comparison: the imbalance reflects the real-life scenario in which non-speakers of a language are far easier to find than speakers of the language, who may not be available at all.

In our comparison with cross-lingual models, the main difficulty is the strength of pre-trained embeddings for the baseline models. As a strong language-independent baseline for existing cross-lingual methods, we trained models on English NER data and evaluated on the target language test data (experiments with related languages showed similar results, and were omitted for space constraints). Our experimental decision to use relatively high-resource languages meant that mBERT models had access to reasonably large amounts of pre-training data (each language was in the top 50 by Wikipedia size), and are therefore unfairly strong. One would expect cross-lingual performance to decrease on lower resource languages (Wu and Dredze, 2020).

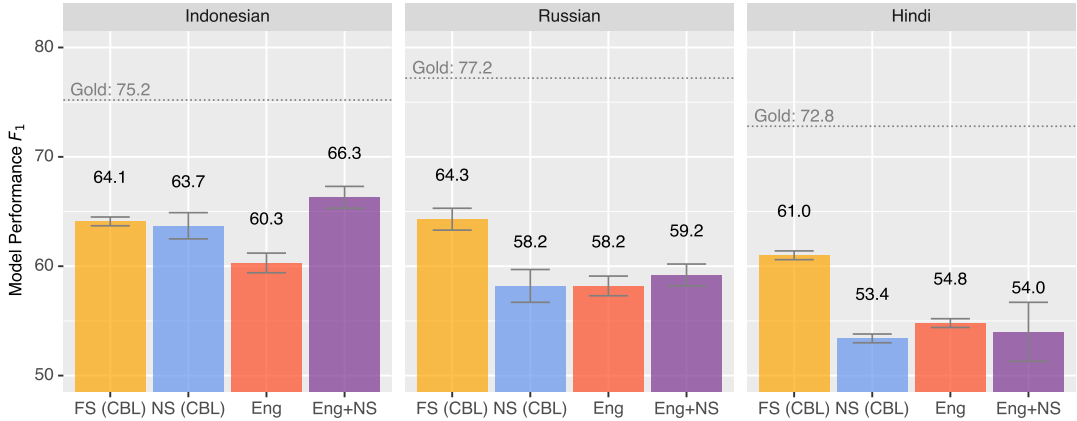

Figure 3: Comparison of models trained on fluent speaker (FS) and non-speaker (NS) annotations to English cross-lingual models, showing comparable or improved performance across all languages. Error bars show one standard deviation calculated over five trials. CBL refers to Constrained Binary Learning. The Eng+NS model is trained on the concatenation of English and NS data. The dashed lines refer to the performance of models trained on the gold annotated training set.

### 4.2 Main Results

Figure 3 summarizes the results of this experiment by providing a comparison of models trained on non-speaker (NS) and fluent speaker (FS) annotations to cross-lingual models. From these results we distill two main takeaways.

**Takeaway 1: NS Annotation Works**

The results of our experiments show that across all languages, non-speaker annotations have produced meaningful results. In Indonesian, NS models are evidently strong and perform at a similar level to models trained on fluent-speaker annotations. This is likely attributable to the high entity overlap between English and Indonesian and limited language-specific information required for successful annotation. In practice, this indicates that 2.5 hours of language exposure was enough for NS annotators to produce annotations with quality sufficient enough to be useful.

The gap between NS and FS model performance widens on other languages, and correlates with an annotation quality drop. This suggests that 2.5 hours are not sufficient to produce NS annotations rivaling FS models (however, as we will see in Takeaway 2, this is sufficient to rival cross-lingual baselines). One reason is that in more difficult languages, annotators need more time to become acquainted with the language, so we could expect more substantial improvements over time. To test this hypothesis, we examined mean annotation quality trends of NS annotators, summarized in Table 5. Across all languages, we see annotators improving over time. For Russian and Hindi in particular, we observe a more overt learning curve indicating that there are more nuances to these languages which must be noticed by annotators over time. This upwards positive trend in annotation quality suggests that the NS results reported here are not the peak results that could be achieved. With additional training and experience, NS annotators can produce stronger results even in more difficult languages.

**Takeaway 2: NS Remains On Par With Cross-Lingual Baselines**

In Figure 3, across all languages performance of models built on NS annotations (blue bars) consistently matches or exceeds the performance of cross-lingual models (red bars). Again, in a low-resource scenario we might expect cross-lingual model performance to drop substantially, so the fact that they are comparable in this situation is encouraging. Additional experiments combining NS and English data (purple bars) shows improvements in Indonesian and Russian, but inconclusive changes in Hindi. Altogether, these results demonstrate that using NS annotations is one of the most effective available ways of building an NER model in a low-resource scenario.

While an unexpected observation shows that FS scores are always 15–20 points below models trained on gold-annotated data, we hypothesize that this difference is mainly attributed to training level and not language ability (Geva et al., 2019).

### 4.3 Low-resource Applications

Although the use of non-speaker annotators has little representation in the research community, there have been several projects that lean on this idea heavily. In the LORELEI evaluations,[2] research groups were tasked with producing NLP tools on truly low-resource languages (including Kinyarwanda, Sinhalese, Ilocano, and Odiya) within a short time frame. A number of new techniques came out of these evaluations, and many groups resorted to using non-speaker annotators (Cheung et al., 2017; Mayhew et al., 2017, 2018, 2019b). In each group, annotators were trained more thoroughly than in the empirical study here, and exhibited a more focused and long-term effort. However, in these projects, the goal was to maximize the final score, not make careful observations of the annotation process. This paper fulfills that need.

## 5 Discussion

While Section 4 showed quantitative outcomes of experimental processes, this section explores the many factors that can contribute to obtaining high quality NS annotations.

**NS Annotation Practices & Strengths** When capitalization is available in the target language, it is a strong indicator for named entities. Analyzing NS annotations over languages with capitalization – Indonesian and Russian – shows that over 90% of annotated tokens are capitalized, a rate similar to what we would expect in English.

For languages with non-Latin scripts – Russian and Hindi – NS annotators often relied on phonetic clues and always annotated on romanized versions of the text. Having access to well-romanized text is critical, as it helps NS annotators make connections between English cognates or previously tagged entities. Some real examples of phonetically recognizable entities from Hindi are:

*paakistaan, biibiisii hindii, baamglaadesh*

A majority of entities tagged in languages with no capitalization are either geo-political entities (i.e. Pakistan, America) or well-known Western names (i.e. Obama, Twitter, BBC). Once an annotator learns a word representation in the target language, they tend to tag every instance as an entity. As a result, we found that NS annotators tend to tag a proportionally less diverse set of entities than FS annotators.

**What makes a good annotator?** Analyzing participant language familiarity and instructional quiz scores shows that neither multilingualism nor initial guideline understanding present a clear predictor for good annotators. Participants who performed best were detail-oriented, patient, and often proactively vocalized their interest in the task or the top annotator award incentive.

One strength of human non-speaker annotators to annotate NER is that, unlike an automatic system, they are able to make inferences over common sense world knowledge. For example, they may use a header to pick out the domain of a document, or use neighboring entities to inform decisions, as in Figure 1, where the presence of *New York* suggests *Central Park* as an entity.

**How does this generalize to other tasks?** Looking to other NLP tasks, it seems clear that NS annotations of conceptually in-depth tasks such as dependency parsing or textual entailment are unlikely to have usable quality. However, for tasks such as part of speech tagging, it could be possible, especially with the help of a tag lexicon and an elementary grammar.

## 6 Conclusion

We demonstrate the effectiveness of using non-speaker annotations as an alternative to cross-lingual methods for building low-resource NER models. A qualitative exploration of the resulting data provides insights about what makes NS annotators so unintuitively successful. One avenue for future exploration is with active learning (Settles, 2009), which has been shown to help in low-resource situations (Chaudhary et al., 2019). Further work may also explore optimal ways to combine NS annotators with FS annotators, should they be available.

---

[2] https://www.nist.gov/itl/iad/mig/lorehlt-evaluations

# 7 Acknowledgement

This work was supported by Contract HR0011-18-2-0052 and Contract HR0011-15-C-0113 with the US Defense Advanced Research Projects Agency (DARPA). Approved for Public Release, Distribution Unlimited. The views expressed are those of the authors and do not reflect the official policy or position of the Department of Defense or the U.S. Government.